\documentclass[letterpaper]{article} 
\usepackage{aaai24}  
\usepackage{times}  
\usepackage{helvet}  
\usepackage{courier}  
\usepackage[hyphens]{url}  
\usepackage{graphicx} 
\usepackage{natbib}  
\usepackage{caption} 
\usepackage{algorithm}
\usepackage{algorithmic}

\usepackage{amssymb}
\usepackage{amsmath}
\usepackage{makecell} 
\usepackage{multirow} 
\usepackage{graphicx} 
\usepackage{bm} 

\usepackage{newfloat}
\usepackage{listings}
\DeclareCaptionStyle{ruled}{labelfont=normalfont,labelsep=colon,strut=off} 
\floatstyle{ruled}
\newfloat{listing}{tb}{lst}{}
\floatname{listing}{Listing}
\title{DeepBranchTracer: A Generally-Applicable Approach to Curvilinear Structure Reconstruction Using Multi-Feature Learning}
\author{
    Chao Liu\textsuperscript{\rm 1,\rm 2},
    Ting Zhao\textsuperscript{\rm 3},
    Nenggan Zheng\textsuperscript{\rm 1, \rm 2, \rm 4, \rm 5}\thanks{Corresponding author.}
}
\affiliations{
    \textsuperscript{\rm 1}Qiushi Academy for Advanced Studies, Zhejiang University, Hangzhou, Zhejiang, China \\
    \textsuperscript{\rm 2}College of Computer Science and Techology, Zhejiang University, Hangzhou, Zhejiang, China \\
    \textsuperscript{\rm 3}Independent Researcher \\
    \textsuperscript{\rm 4}State Key Lab of Brain-Machine Intelligence, Zhejiang University, Hangzhou, Zhejiang, China \\
    \textsuperscript{\rm 5}CCAI by MOE and Zhejiang Provincial Government(ZJU), Hangzhou, Zhejiang, China
    
    supermeliu@zju.edu.cn, tingzhao@gmail.com, zng@cs.zju.edu.cn
}

\usepackage{bibentry}

\begin{document}

\maketitle

\begin{abstract}
    Curvilinear structures, which include line-like continuous objects, are fundamental geometrical elements in image-based applications. Reconstructing these structures from images constitutes a pivotal research area in computer vision. However, the complex topology and ambiguous image evidence render this process a challenging task. In this paper, we introduce DeepBranchTracer, a novel method that learns both external image features and internal geometric characteristics to reconstruct curvilinear structures. Firstly, we formulate the curvilinear structures extraction as a geometric attribute estimation problem. Then, a curvilinear structure feature learning network is designed to extract essential branch attributes, including the image features of centerline and boundary, and the geometric features of direction and radius. Finally, utilizing a multi-feature fusion tracing strategy, our model iteratively traces the entire branch by integrating the extracted image and geometric features. We extensively evaluated our model on both 2D and 3D datasets, demonstrating its superior performance over existing segmentation and reconstruction methods in terms of accuracy and continuity.
    
\end{abstract}

\section{Introduction}

Curvilinear structure is a line-like, elongated object with varying intensities compared to their surroundings. The extraction of curvilinear structures aims to quantitatively identify these objects from image data, facilitating statistical analyses of branch connections and distribution. This represents a pivotal research area in computer vision, particularly for neuron branches and blood vessels in medical images \cite{meijering2010neuron, soomro2019deep}, and roads in aerial images \cite{chen2022road}. However, despite decades of research efforts dedicated to automating this process, extracting curvilinear structures in occlusion and ambiguous topology images still remains a challenging task.

\begin{figure}[t]
    \centering
    \includegraphics[width=0.46\textwidth]{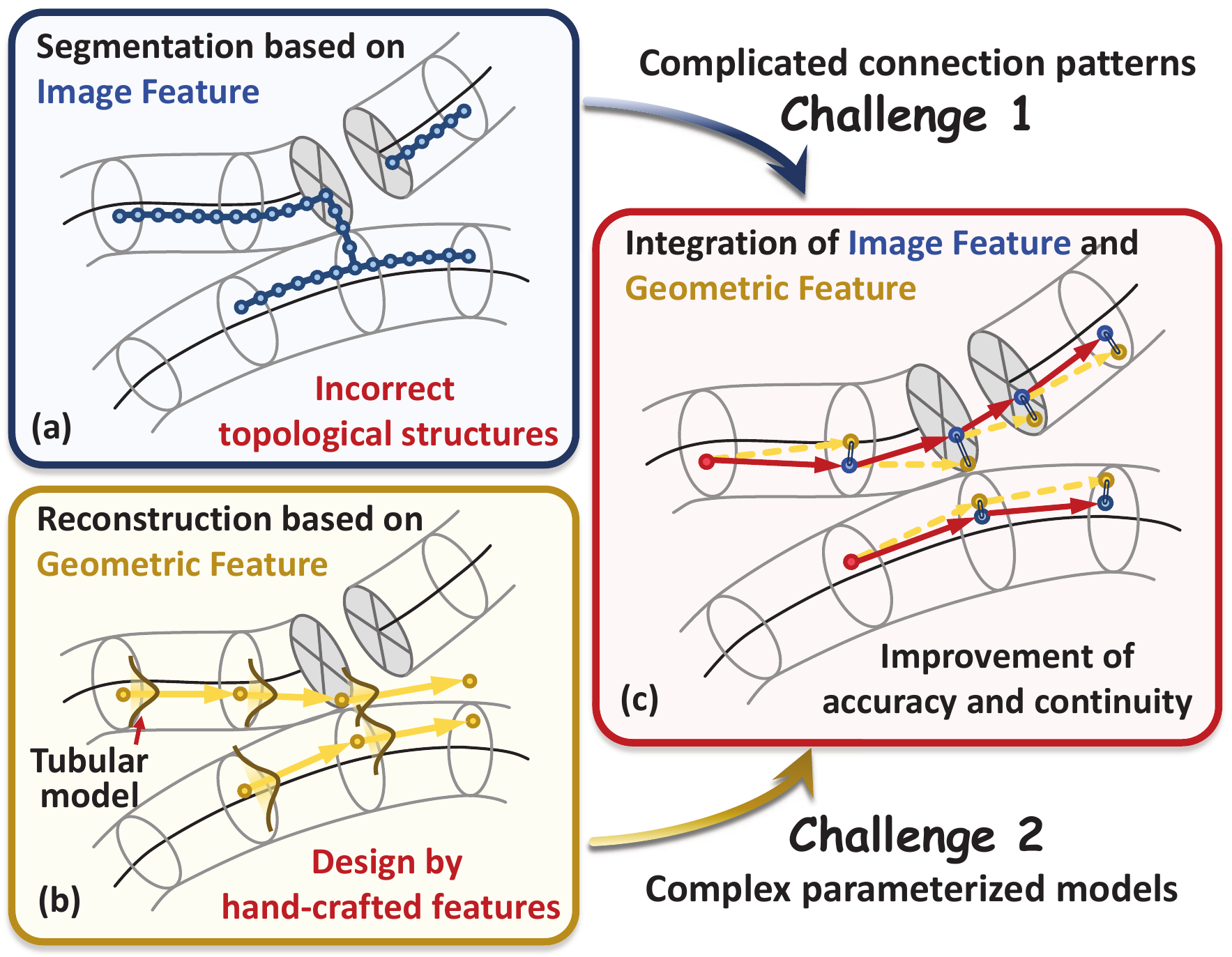}
	\caption{Motivation. Segmentation methods focus on learning image features but tend to misinterpret two adjacent branches as an entity. Reconstruction methods capture the geometric features of branches but rely on complex parameterized models. The integration of these two features offers the potential to improve the continuity of results.}
	\label{firstfig}
\end{figure}

Over the past few years, curvilinear structure extraction has always been addressed as a segmentation-based problem. As an essential segmentation tool, convolutional neural networks (CNNs) can automatically extract image features from diverse images \cite{li2017deep, long2015fully_4}. However, unlike typical segmentation tasks, curvilinear structure extraction places heightened emphasis on the continuity and connectivity of the target, rather than just its contour. Several studies have incorporated the prior structural constraints into the network architectures or loss functions to enhance their effectiveness \cite{qi2023dynamic, zhou2021split, liu2022using}. Nevertheless, a foundational challenge remains across these segmentation-based methods. When confronted with two adjacent or tangled branches, segmentation-based methods often tend to misinterpret them as a single entity, and will constitute a crucial topological error, as depicted in Fig.~\ref{firstfig}(a).



In response to these challenges, reconstruction-based methods have emerged to explicitly model the topological representation of curvilinear structures, as shown in Fig.~\ref{firstfig}(b). These methods encompass both external image features and internal geometric features, and can be categorized into two groups based on their reconstruction processes: \textbf{global-first} approaches and \textbf{local-first} approaches. 

Global-first approaches, as a variant of segmentation-based methods, extract nodes or segments from the whole image perspective to produce a skeleton. These top-down approaches, such as graph optimization \cite{turetken2016reconstructing}, skeletonization \cite{yuan2009mdl} and signal enhancement \cite{yang2021structure}, can efficiently explore weak branches by merging the potential edges into networks. However, a major drawback of these approaches is that they have an increased likelihood of generating \emph{spurious branches} due to mistaken seed points from noise. Therefore, a post-processing pruning algorithm or labor-intensive human correction is necessary to generate the definitive reconstruction \cite{peng2011proof}. 

Local-first approaches start from a reasonable seed point to explore the potential branches. As a conservative strategy, these approaches prioritize local information and cease execution when image signals become indistinct, thereby avoiding the error of spurious branches. Thus, local-first approaches become a promising way for addressing complex topology challenges in large-scale image applications, such as neuron tracing or city road extraction. Nonetheless, these approaches often rely on geometric models designed with hand-crafted features, such as tubular template \cite{zhao2011automated} and ray-shooting model \cite{ming2013rapid}, and the incorporation of these models as modules into a deep network for training is challenging. Local-first approaches also place significant reliance on the accuracy of the global pre-segmentation mask to decide whether to continue tracing, otherwise the weak branches could be omitted in the results.


\begin{figure*}[htb]
	\centering
	\includegraphics[width=0.98\textwidth]{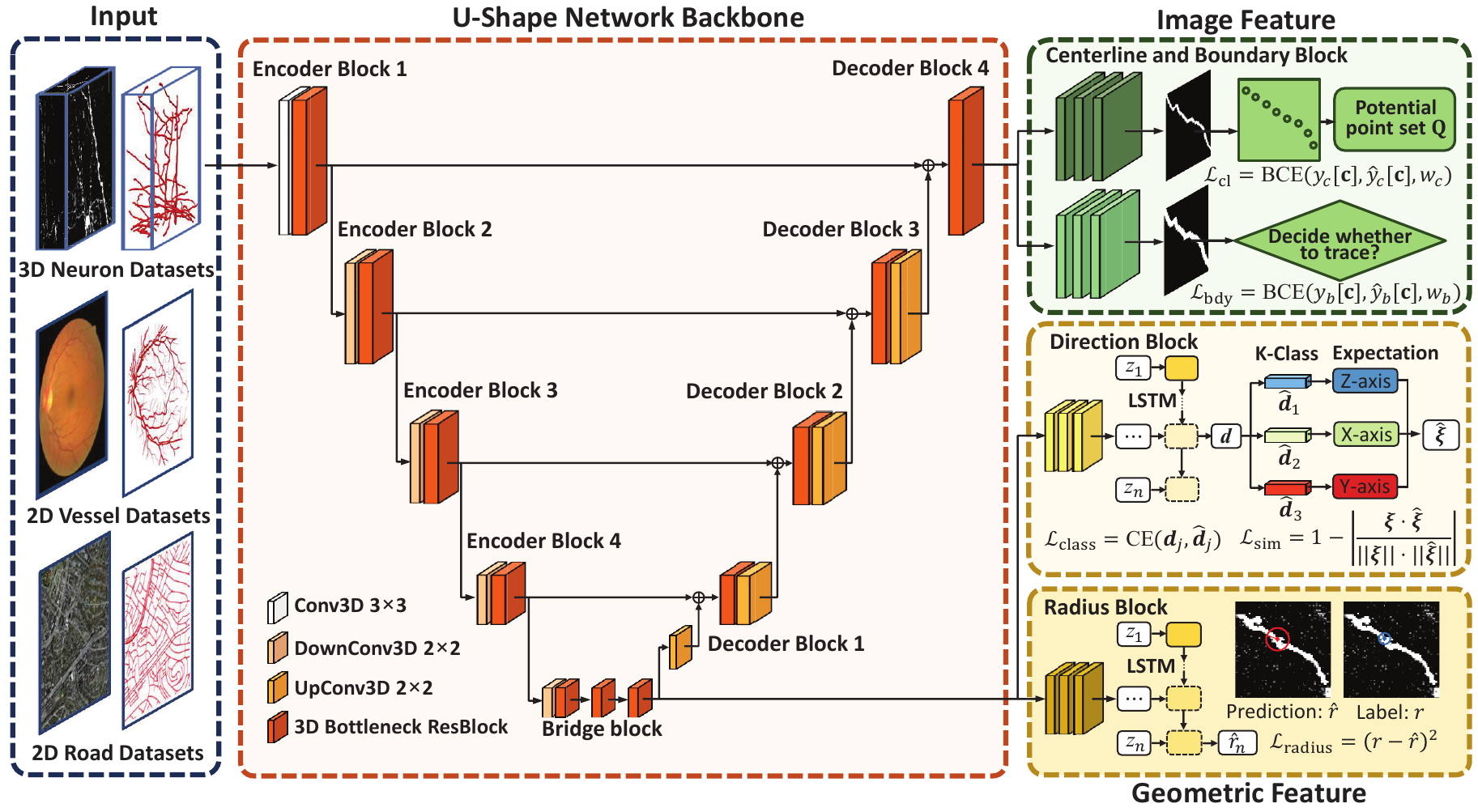}
	\caption{Architecture of the CSFL network. The network integrates four sub-task blocks in a U-shaped network to extract both geometric features (including radius and direction blocks) and image features (including centerline and boundary blocks). }
	\label{framework_1}
\end{figure*}

To overcome the above limitations of the local-first approaches, we proposed a novel reconstruction model called DeepBranchTracer, involving a curvilinear structure feature learning network (CSFL) and a multi-feature fusion tracing strategy (MFT). Firstly, to address the challenge of integrating geometric models into a deep network, we formulate a mathematical framework for describing curvilinear structures and investigate the geometric parameters that necessitate optimization. Secondly, to expand the interest region of local-first approaches, we propose a U-shaped network with four subtasks blocks to learn both the external image feature and internal geometric continuous feature in an integrated CSFL network. For the image features, the centerline serves to fine-tune the branch positions, while the boundary determines whether to stop a tracing process. For the geometric features, a sequence learning module is designed for direction and radius estimation blocks to capture smoothness and continuity of the branch signals. Finally, we developed an MFT strategy to explore the curvilinear structure iteratively based on the branch multiple features extracted above. 

Our main contributions can be summarized as follows:
\begin{itemize}
\item We formulate the problem of curvilinear structure extraction as an explicit geometric parameter estimation task for deep networks, eliminating the need for complex geometric parametric models based on hand-craft features.
\item DeepBranchTracer involves the CSFL network to learn both internal geometric feature and external image feature of branches, and combines the MFT strategy, enabling an iterative tracing process for finalizing the reconstruction.
\item To our best knowledge, DeepBranchTracer is the first work for curvilinear structure reconstruction for both 2D and 3D images. With extensive experiments on five various datasets, our model has achieved better accuracy and continuity compared to other state-of-the-art (SOTA) segmentation and reconstruction methods.
\end{itemize}

\section{Related Work}

\subsection{Segmentation-based Methods}

Various segmentation-based methods for curvilinear structure have been proposed to achieve better results by adjusting model backbones or feature learning modules. As a prominent segmentation architecture, U-Net \cite{ronneberger2015u} has inspired a diverse of variant network models for curvilinear structure segmentation, including the integration of ResBlock \cite{milletari2016v}, feature aggregation \cite{liu2022full}, attention mechanisms \cite{schlemper2019attention}, among others. To improve the continuity of the results, the deformable convolution \cite{dai2017deformable} and its derivatives \cite{yang2022dcu, qi2023dynamic} were proposed to permit the network to dynamically adjust the receptive field, facilitating the learning of geometric object features.  Additionally, SGCN \cite{zhou2021split} employed a split depth-wise separable graph convolutional network to capture global contextual information in channel and spatial features. Nonetheless, a foundational challenge remains across these segmentation-based methods. When input image evidence contains fragmented or obstructed branches, networks struggle to learn structural features effectively, and require a post-processing to repair the broken segmentation results.






\subsection{Reconstruction-based Methods}

The reconstruction-based methods can be divided into two categories. As an example of the global-first approaches, \cite{turetken2016reconstructing} introduced an integer programming technique to identifying the optimal subset of paths, subject to topological constraints that eliminate implausible solutions. On the basis of this, \cite{mosinska2019joint} trained a deep network to segment and evaluate the potential paths as a data-driven method. In contrast, local-first approaches start from a set of seed points and employ a greedy tracing methodology guided by geometric template fitting score. These methods are effective in certain applications but rely on manually crafted features, such as voxel scooping \cite{rodriguez2009three}, ray shooting \cite{ming2013rapid}, and tubular fitting \cite{zhao2011automated}, making them difficult to choose tuning parameters when faced with different datasets. The end-to-end reconstruction methods SPE \cite{chen2021deep} and RoadTracer \cite{bastani2018roadtracer} have also been proposed to iteratively estimate tracing directions. SPE employs a spherical core to transform the 3D stacks of neuron images into 2D patches and also leads to the loss of texture information in the original images, limiting its applicability to other fields beyond microscope grayscale images. RoadTracer constructs a CNN-based model that iteratively builds a road network graph using direction estimation and a decision function. However, it lacks a direction repair module, which will alway cause an early stop in the tracing process in case of an unexpected prediction error. 

Compared with earlier tracing methods, our model eliminates the need for complex parameterized geometric models, seamlessly integrates image features and geometric features to enhance the tracing capabilities, particularly for elongated branches.


\section{Method}

In this section, we formulate the problem of curvilinear structure reconstruction as the task of geometric parameters estimation, including radius, direction and position. Subsequently, we introduce our DeepBranchTracer model, which leverages a CSFL network and four sub-task blocks to acquire insights from both image and geometric features. Finally, we detail the incorporation of learned branch attributes into the tracing process through the MFT strategy.

\subsection{Problem Formulation}
A curvilinear structure can be represented as a continuous solid whose axis is a 3-D space curve. This curve is a set of points $C \! \subset \! \mathbb{R}^3$ with the property that for every $\mathbf{c} \in C$ there exists a smooth function $\{ \mathbf{c}(t) \!=\! (x(t),y(t),z(t))^T, t\!\in\!(a, b) \}$, where $\mathbf{c}(t)$ is the centerline of the curvilinear structure. Each point also has a property of radius $r(t)$ to form a circular cross section. Along the centerline, we can integrate these circular cross sections and form a solid curvilinear structure $S(t)$ to fit the branch object in image $I$. To facilitate the calculation, we approximated the continuous centerline as $m$ pieces of segments, and the curvilinear structure $S(t)$ can be represented by the discrete cylinders $S_i(t)$:
\begin{equation}
    S(t) = \sum ^{m}_{i=1} S_i(t),
\end{equation}
and each cylinder can be represented as:
\begin{equation}
    \! S_i(t) \! \approx \!
    \left\{\!
    \begin{array}{cl}
        \!(1\!-\!\gamma(t))\mathbf{c}(t_i) \!+ \!\gamma(t)\mathbf{c}(t_{i+1}) \! & \! \multirow{2}{*}{\!, \!\text{if $t \!\in \!(t_{i}, \!t_{i+1})$}} \vspace{0.5ex}\\ 
        + \bm{R}(t_i) \mathbf{D}(t_i) \mathbf{O}(u) &  \vspace{1ex}\\
        \! 0 \! & \!  \text{\!, \!otherwise} \\
    \end{array}
    \right. \!,
    \label{eq:s(t)}
\end{equation}
where
\begin{equation}
    \gamma(t) = \frac{t-t_i}{t_{i+1} - t_i},
\end{equation}
where $\bm{R}(t_i)\!=\!\varphi \cdot r(t_i)\!$ represents the space within the cylinder radius, and $\varphi \!\in\! (0, 1]$. $\mathbf{D}(t_i)\!=\!(\pmb{\nu}(t_i) , \pmb{\zeta}(t_i))$ are two orthogonal vectors to form the normal plane of the centerline with reference to the unit direction vector $\pmb{\xi}(t_i)$, and can be obtained based on the Frenet frame, which is elaborated in the Appendix. $\mathbf{O}(u)\!=\!(\cos u, \sin u)^T$ is the parametric equation of a circle, and $u \!\in\! (0,2\pi]$.



Thus, the problem of accurately reconstructing a curvilinear structure can be viewed as an iterative process involving the estimation of the position point $\mathbf{c}(t_i)$, its radius $r(t_i)$, and direction $\pmb{\xi}(t_i)$ to determine the subsequent point $\mathbf{c}(t_{i+1})$: 
\begin{equation}
    \mathbf{c}(t_{i+1}) = \mathbf{c}(t_{i}) + r(t_i) \pmb{\xi}(t_i).
    \label{eq:trace}
\end{equation}
As a constraint condition, all the point $\mathbf{c}(t_{i})$ should adhere to the condition $\mathbf{c}(t_{i}) \! \in \! S(t)$ to prevent deviation from the centerline. Finally, this process yields the ultimate solid $S(t)$ as the reconstruction result. Here, we propose the overall loss function and look for the suitable $r^*$, $\pmb{\xi}^*$ and $\mathbf{c}^*$:
\begin{equation}
    \begin{aligned}
    \min  \sum^{m}_{i=1} & ( \mathcal{L}_{\text{geo}}(r(t_i), \pmb{\xi}(t_i), I[\mathbf{c}(t_{i-l:i})]) \\
    & + \mathcal{L}_{\text{img}}(y_b[\mathbf{c}(t_i)], y_c[\mathbf{c}(t_i)], I[\mathbf{c}(t_i)]) ) ,\\
    \end{aligned} 
\end{equation}
where $I[\mathbf{c}(t_i)]$ is the image at the point $\mathbf{c}(t_i)$, $I[\mathbf{c}(t_{i-l:i})]$ represents the the image patches from $l$ historical points, and $y_b$, $y_c$ are the segmentation labels of image. $\mathcal{L}_{\text{geo}}$ represents the geometry loss associated with radius and direction estimation, while $\mathcal{L}_{\text{img}}$ signifies the image loss employed to incorporate image features during the tracing process. We will describe these two loss terms in more detail below.

\subsection{Curvilinear Structure Feature Learning Network}

To estimate the above branch properties, we employ the CSFL network to extract the image and geometric features using four sub-task blocks: centerline and boundary blocks, direction and radius blocks, as depicted in Figure \ref{framework_1}.

\subsubsection{U-shaped Backbone.}
The U-shaped backbone comprises three main components: an encoder, a bridge, and a decoder. The encoder is implemented as a 4-level bottleneck ResBlock, while the bridge exhibits a broader 3-level block compared to the original U-Net model. This enhanced configuration allows for the extraction of more comprehensive high-level semantic information, and can be represented as $f_{\phi_E} \!:\! I[\mathbf{c}(t_i)] \!\to\! z_{e}(t_i)$. The output of the bridge serves as the input for both the direction and radius block, enhancing the provision of geometric features. The decoder exhibits an analogous architecture to the encoder, with the goal of upsampling the encoded features to match the original resolution of the input image, and can be represented as $f_{\phi_D} \!:\! z_{e}(t_i) \!\to\! z_{d}(t_i)$. It generates inputs for the centerline and boundary blocks.

\subsubsection{Geometric Feature.}

In the previous work, several methods \cite{chen2021deep, bastani2018roadtracer} have employed CNNs to predict geometric features within a local context. However, it is widely observed that points along a continuous branch often manifest similar direction and radius values in comparison to their neighboring points. To capture this natural attribute, we incorporate the sequential learning module into the radius and direction blocks, facilitating the capture of branch smoothness and continuity. The feature maps derived from historical continuous points are fed as input to predict the geometric features of the successor point.

We have devised the direction block using an LSTM module, serving as the mapping function $f_{\theta_d} \!:\! z_e(t_{i-l:i}) \!\to\! d(t_i)$. The direction block is composed of two losses, which is the classification loss of bins denoted as $\mathcal{L}_{\text{class}}$ and the similarity loss of direction referred to as $\mathcal{L}_{\text{sim}}$. The task of $K$-class classification for direction estimation is designed to ascertain the bin to which the vector component belongs:
\begin{equation}
    \mathcal{L}_{\text{class}} = \frac{1}{J} \sum^{J}_{j=1} \text{CE}(\bm{d}_{j}(t_i),\hat{\bm{d}}_{j}(t_i)),
\end{equation}
where $J$ is the number of vector component. In 3D space, $J$ equals 3, and $\bm{d}(t_i)$ comprises components along the $X$, $Y$, and $Z$ axes, denoted as $[\bm{d}_{x}(t_i), \bm{d}_{y}(t_i), \bm{d}_{z}(t_i)]$.

However, the adjacent bins of the classification are inter-connected in the direction estimation task. For instance, when the direction label is [0, 1, 0, 0, 0], the predicted probability of [0, 0, 0.9, 0, 0] is expected to be superior to the probability of [0, 0, 0, 0.9, 0], even though they have the same classification loss. Therefore, from a geometric perspective, in order to obtain the fine-grained direction predictions, we introduce a cosine similarity loss that computes the disparity between the two direction vectors:
\begin{equation}
\mathcal{L}_{\text{sim}} = 1- \left | \frac{\pmb{\xi}(t_i) \cdot \hat{\pmb{\xi}}(t_i)}{||\pmb{\xi}(t_i)||\cdot||\hat{\pmb{\xi}}(t_i)||} \right |,
\end{equation}
and for each vector component of $\hat{\pmb{\xi}}(t_i)$:
\begin{equation}
\hat{\pmb{\xi}}_j(t_i) = \sum^{K}_{k=1} \text{softmax} (\hat{\bm{d}}_{j}(t_i))_k \cdot A_k
\end{equation}
where $K$ is the number of classes, and $A_k$ is the expectation of each bin. The direction $\hat{\pmb{\xi}}(t_i)$ is computed by taking the expectation of the binned output for each vector component. Specifically, a direction can be represented as the positive $\pmb{\xi}$ and negative $-\pmb{\xi}$ vectors of the same value but diff sign. When the cosine similarity between the predicted and ground truth vectors equals 1 or -1, the $\mathcal{L}_{\text{sim}}$ will be 0, signifying a higher direction similarity. The final loss of direction block represented as:
\begin{equation}
    \mathcal{L}_{\text{direction}}= \mathcal{L}_{\text{class}} + \mathcal{L}_{\text{sim}}.
\end{equation}

Similarly, the radius block utilizes the same sequential learning module and is designed as the mapping of $f_{\theta_r} \!:\! z_e(t_{i-l:i}) \!\to\! \hat{r}(t_i)$. To train this module, the mean-squared error loss is employed:
\begin{equation}
\mathcal{L}_{\text{radius}} = ( r(t_i) - \hat{r}(t_i) )^2.
\end{equation}

Combined the direction block and the radius block, the final geometry loss is integrated as:
\begin{equation}
    \mathcal{L}_{\text{geo}}= \lambda_{d} \mathcal{L}_{\text{direction}} + \lambda_{r} \mathcal{L}_{\text{radius}},
    \label{eq:loss_geo}
\end{equation}
where $\lambda_{d}$ and $\lambda_{r}$ are the weights to balance these two losses.

\subsubsection{Image Feature.} The purpose of centerline extraction is to provide the potential successor points based on the image features. Here, we consider the problem of centerline extraction as a binary segmentation task. However, the centerline of curvilinear structure is often thin and line-like, leading to an imbalance of positive and negative data in the training dataset. To address this issue, we employ a weighted binary cross entropy loss function, which assigns a higher cost to the foreground branch in order to improve the performance. The centerline loss $\mathcal{L}_{\text{cl}}$ is formulated as:
\begin{equation}
    \mathcal{L}_{\text{cl}} = \text{BCE}(y_c[\mathbf{c}(t_i)], \hat{y}_c[\mathbf{c}(t_i)], w_c),
    \label{eq:loss_cl}
\end{equation}
where $y_c[\mathbf{c}(t_i)]$ and $\hat{y}_c[\mathbf{c}(t_i)]$ indicate the ground truth and probability map of centerline, $w_c$ is the weight for foreground of centerline. Based on the Eq.~\eqref{eq:s(t)}, we set the radius $r$ of all points to 1 to generate the centerline label.

The boundary block determines whether to continue tracing in the next step. Previous researches \cite{chen2021deep, bastani2018roadtracer} employed a fully connected layer classifier to ascertain the probability of the present location belonging to either the foreground \emph{branch} or background \emph{noise}. Nonetheless, this approach consolidates all feature maps into a singular vector, resulting in the loss of position information within the image. Here, we approach the tracing decision as a boundary segmentation task:
\begin{equation}
    \mathcal{L}_{\text{bdy}} = \text{BCE}(y_b[\mathbf{c}(t_i)], \hat{y}_b[\mathbf{c}(t_i)], w_b),
    \label{eq:loss_bdy}
\end{equation}
where $\hat{y}_b[\mathbf{c}(t_i)]$ is the probability map of the boundary, $w_b$ is the weight for foreground of boundary. We set the radius $r$ of all points to $r_b$ to generate the boundary label. When $r_b$ is a large value, the tracing algorithm tends to adopt an aggressive strategy; conversely, it becomes conservative.

The overall loss function of image feature can be defined as follows:
\begin{equation}
    \mathcal{L}_{\text{img}} = \lambda_{c} \mathcal{L}_{\text{cl}} + \lambda_{b} \mathcal{L}_{\text{bdy}},
    \label{eq:loss_img}
\end{equation}
where $\lambda_{c}$ and $\lambda_{b}$ are the weights to balance the two losses.

\subsection{Multi-Feature Fusion Tracing Strategy}

\begin{figure}[t]
    \centering
    \includegraphics[width=0.46\textwidth]{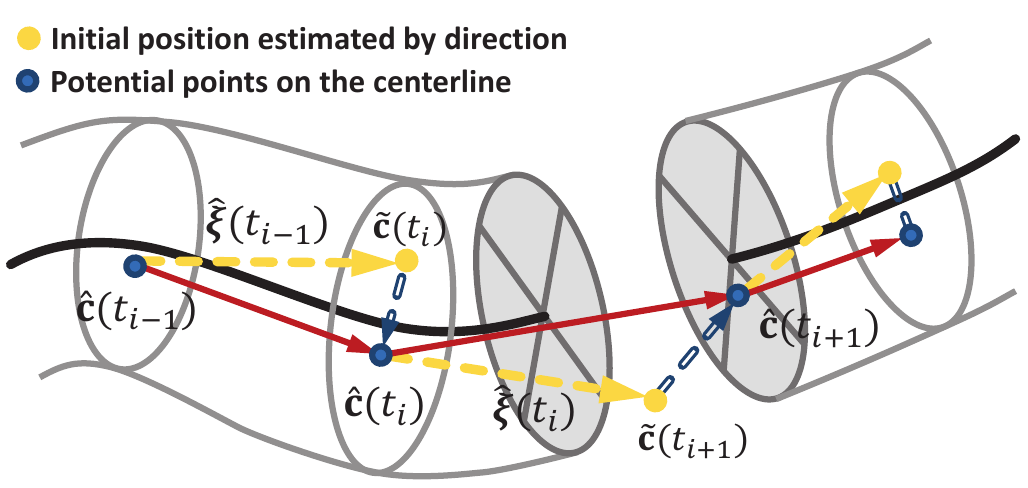}
	\caption{MFT strategy. Step 1: Find an initial point $\tilde{\mathbf{c}}(t_i)$ based on the geometric features of direction and radius. Step 2: From the point set $\mathbf{Q}$ of the surrounding centerline, locate the most potential point $\hat{\mathbf{c}}(t_i)$ to adjust the position. Step 3: Determine whether to trace the successor point based on the boundary probability $\hat{y}_b(\hat{\mathbf{c}}(t_{i}))$.
    }
	\label{tracingstrategy}
\end{figure}

Our model utilizes an MFT strategy, combining both internal geometric features and external image features to trace the branch iteratively, as shown in Fig.~\ref{tracingstrategy}.

Starting from a seed point $\mathbf{c}(t_{i})$, the module traces the branch in two steps. For the first step, we estimate the radius $\hat{r}(t_{i})$ and direction $\hat{\pmb{\xi}}(t_{i})$ of the current point to find an initial successor point $\tilde{\mathbf{c}}(t_{i+1})$ based on the Eq.~\eqref{eq:trace}. However, due to low quality and ambiguity of image, the position of the $\tilde{\mathbf{c}}(t_{i+1})$ is always low confidence. If our model follows this decision, it may deviate from the centerline. Thus, in the second step, our goal is to enhance the accuracy of the branches by incorporating the centerline feature. This involves identifying the closest point $\hat{\mathbf{c}}(t_{i+1})$ from the potential successor point set $\mathbf{Q}$ to current point $\tilde{\mathbf{c}}(t_{i+1})$ as the final decision:
\begin{equation}
    \hat{\mathbf{c}}(t_{i+1}) = \mathop{\arg\min}\limits_{\mathbf{c}(q) \in \left \{ \mathbf{Q} |\hat{y}_c[\hat{\mathbf{c}}(t_{i})] > T_c \right \}} \left \| \tilde{\mathbf{c}}(t_{i+1}) - \mathbf{c}(q) \right \| ^2,
\end{equation}
where $T_c$ is the threshold of the probability map $\hat{y}_c[\hat{\mathbf{c}}(t_{i})]$. Points within the centerline probability map exhibiting values exceeding $T_c$ are included within the potential successor point set $\mathbf{Q}$.

Furthermore, we utilize two stopping criteria for the tracing process. The first criterion involves checking if the boundary probability of the successor point $\hat{\mathbf{c}}(t_{i+1})$ falls below a threshold $T_b$. The second criterion is triggered when the successor point intersects with a previously traced region. The detail of the tracing strategy is laid out in Algorithm \ref{alg:algorithm}.

\begin{algorithm}[t]
    \caption{Multi-Feature Fusion Tracing Strategy}
    \label{alg:algorithm}
    \textbf{Input}: Seed point position: $\mathbf{c}(t_1)$, Input image $I$\\
    \textbf{Output}: Reconstructed branch: $B$
    \begin{algorithmic}[1] 
        \STATE Initialize $B$ as the empty point set $B \gets \emptyset$, tracing action $a = \text{continue}$, $i \gets 1$

        \WHILE{$a = \text{continue}$}
        \STATE // \textit{Predict branch and direction parameters.} \\
        \STATE $ \hat{y}_c[\mathbf{c}(t_i)], \hat{y}_b[\mathbf{c}(t_i)], \hat{r}(t_{i}), \hat{\pmb{\xi}}(t_{i}) \gets \text{Network}(I[\mathbf{c}(t_i)])$ \\

        \STATE // \textit{Estimate initial point by direction.}
        \STATE  $\tilde{\mathbf{c}}(t_{i+1}) = \mathbf{c}(t_{i}) + \hat{r}(t_i)  \hat{\pmb{\xi}}(t_i) $ \\

        \STATE // \textit{Adjust position based on centerline.}
        \STATE $\hat{\mathbf{c}}(t_{i+1}) = \mathop{\arg\min}\limits_{\mathbf{c}(q) \in \left \{ \mathbf{Q} |\hat{y}_c[\mathbf{c}(t_{i})] > T_c \right \}} \left \| \tilde{\mathbf{c}}(t_{i+1}) - \mathbf{c}(q) \right \| ^2$ \\
        \IF {$\hat{y}_b(\hat{\mathbf{c}}(t_{i+1})) > T_b $ and $\hat{\mathbf{c}}(t_{i+1})$ is not in $B$}
        \STATE $\mathbf{c}(t_{i+1}) \gets \hat{\mathbf{c}}(t_{i+1}) $
        \STATE $B \gets B \cup  \{\mathbf{c}(t_{i+1})\}$
        \STATE $i \gets i+1$
        \ELSE
        \STATE $a = \text{stop}$
        \ENDIF
        \ENDWHILE
        \STATE \textbf{return} $B$
    \end{algorithmic}
\end{algorithm}




\section{Experiments}
We carried out ablation experiments of our model DeepBranchTracer in order to evaluate the effects of different model settings, including the image feature, the geometric feature, and the LSTM sequence learning module. We also quantified the performance of our approach on both 2D and 3D datasets. The results show that our proposed model achieved better results compared with the SOTA segmentation-based and reconstruction-based models.

\subsection{Dataset}

We conducted experiments on five popular datasets contain 2D and 3D images, consisting of Massachusetts Roads Dataset \cite{mnih2013machine}, DRIVE \cite{staal2004ridge} and CHASE\_DB1 \cite{carballal2018automatic} retina datasets, 3D DIADEM-CCF \cite{brown2011diadem} and fMOST-VTA \cite{li2010micro, gong2013continuously} neuron datasets. Specifically, the fMOST-VTA dataset was extracted from the Ventral Tegmental Area (VTA) of a mouse brain imaged by fMOST with the size of 9216 $\times$ 14336 $\times$ 24576 voxels and has a spatial resolution of 0.325 $\times$ 0.325 $\times$ 1 $\mu m^3$. We pre-processed it into the 16-bit gray image stacks with the size of 512 $\times$ 512 $\times$ 512 voxels. With the aid of neuTube open-source software \cite{feng2015neutube}, a semi-automatic neuron reconstruction software, 25 image stacks were manually annotated by experts to serve as the ground truth. Among them, 20 were used for training and 5 for testing.


\begin{figure}[t]
	\centering
    \includegraphics[width=0.46\textwidth]{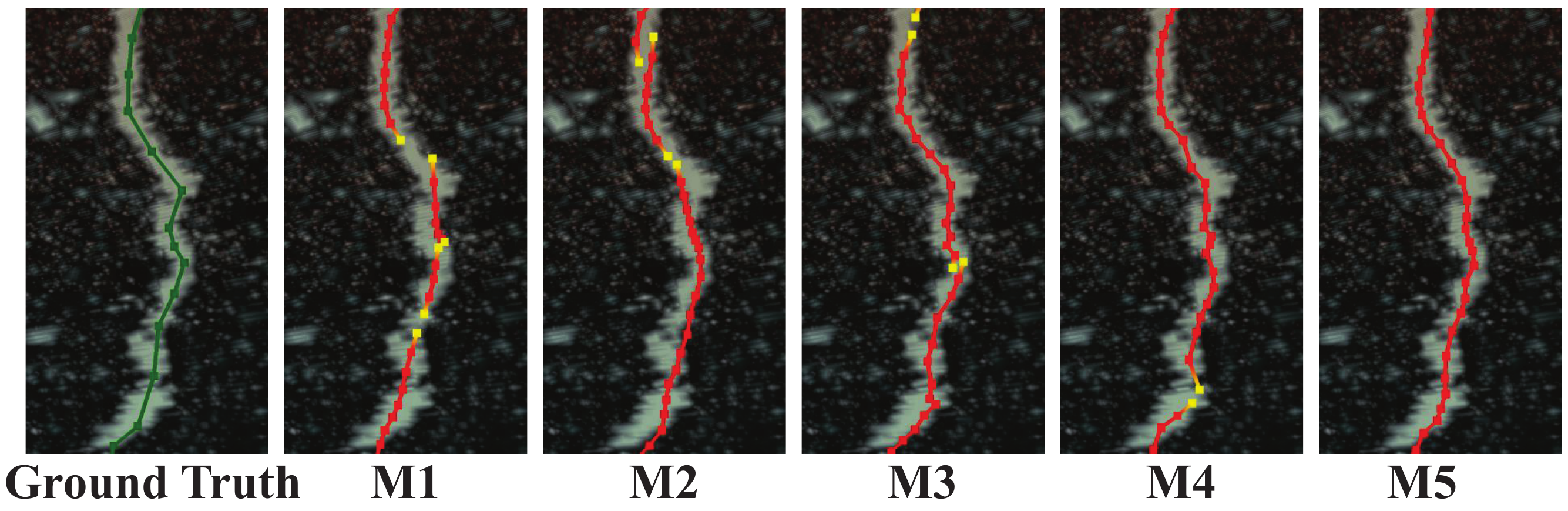}
	\caption{From left to right, with the integration of centerline and geometric features, the reconstruction branch of M5 achieves improved smoothness and continuity. The yellow dots represent the termination points of the broken branches.}
	\label{ablstudy_neuron}
\end{figure}

\begin{table}[t] 
    \centering
    \small
    \setlength{\tabcolsep}{2pt}
    \begin{tabular}{c!{\vrule width1pt}ccc!{\vrule width1pt}ccccc}
    \Xhline{1.2pt}
        & $\mathcal{L}_{\text{cl}}$ &  $\mathcal{L}_{\text{geo}}$ & LSTM & PE    & RE    & SSD-F1  & Length-F1  & ABL    \\
    \Xhline{0.8pt}
    M1 &              & \checkmark   &             & 1.879 & 0.281 & 0.502  & 0.704     & 48.07  \\
    M2 &              & \checkmark   & \checkmark  & 1.816 & 0.270 & 0.519  & 0.701     & 57.04  \\
    M3 & \checkmark   &              &             & 0.835 & 0.264 & 0.845  & 0.893     & 128.65 \\
    M4 & \checkmark   & \checkmark   &             & 1.021 & 0.265 & 0.893  & 0.890     & 167.51 \\
    M5 & \checkmark   & \checkmark   & \checkmark  & \textbf{0.775} & \textbf{0.262} & \textbf{0.910}  & \textbf{0.911}     & \textbf{198.33} \\
    \Xhline{1.2pt}         
    \end{tabular}
    \caption{Ablation study for geometric and image feature learning modules based on fMOST-VTA dataset.}
    \label{Ablation_table}
\end{table}

\subsection{Implementation Details}
During the training process, all the 2D images are split into small patches of size 64  $\times$ 64 with the batch size of 64, and all the 3D images are split into small patches of size 16  $\times$ 64  $\times$ 64 with the batch size of 16. The hyper-parameters in Eq.~\eqref{eq:loss_geo} and Eq.~\eqref{eq:loss_img} are set to $\lambda_d=1$, $\lambda_r=100$, $\lambda_c=1$, and $\lambda_b=1$. The weights of binary cross entropy loss function Eq.~\eqref{eq:loss_cl} and Eq.~\eqref{eq:loss_bdy} are set to $w_c=w_b=0.9$. The thresholds in the MFT strategy are set to $T_c=0.5$ and $T_b=0.5$. During the training process, the network is optimized by minimizing the centerline loss $\lambda_c$ and the boundary loss $\lambda_b$. After the network converges, we freeze the parameters of U-shaped backbone, centerline block, and boundary block, and optimize the other two direction and radius blocks by minimizing the loss $\lambda_d$ and $\lambda_r$. The network architecture is implemented in PyTorch 1.7.0 and MindSpore 1.7.0 \cite{huawei2022deep}. The hardware configuration has 64 GB of memory, and the CPU is an Intel Xeon E5-2680 v3 and four NVIDIA GeForce RTX2080ti GPUs. With this server configuration, we took about 12 hours to train the model. The code of this work is available at \emph{https://github.com/CSDLLab/DeepBranchTracer}.

\subsection{Evaluation Metrics} 
Curvilinear structure is an elongated object, so the evaluation metric of curvilinear structure extraction should be concerned with the accuracy of its geometry rather than the classical segmentation effect. Here, we chose seven topology metrics to assess the performance of reconstruction methods. SSD-F1 metric is proposed in \cite{peng2010v3d}, aiming to measure how many nodes are matched in the reconstructed branch and the ground truth branch. The position error (PE) and radius error metric (RE) are Euclidean distances that calculate the position and radius of the matching node between two branches. Length metric \cite{wang2011broadly, zhang2022pyneval} (including Precision, Recall and F1-score) can be viewed as a variant of the SSD-F1 metric in terms of it tries to measure how many edges are matched. Precision indicates how accurately the predicted branch locations are. Recall captures the overall proportion of the ground truth that is effectively covered by prediction branches. We also calculated the average length of the branches (ABL), hoping that there will be more long and intact branches in the reconstruction results. 

\begin{figure}[]
	\centering
    \includegraphics[width=0.46\textwidth]{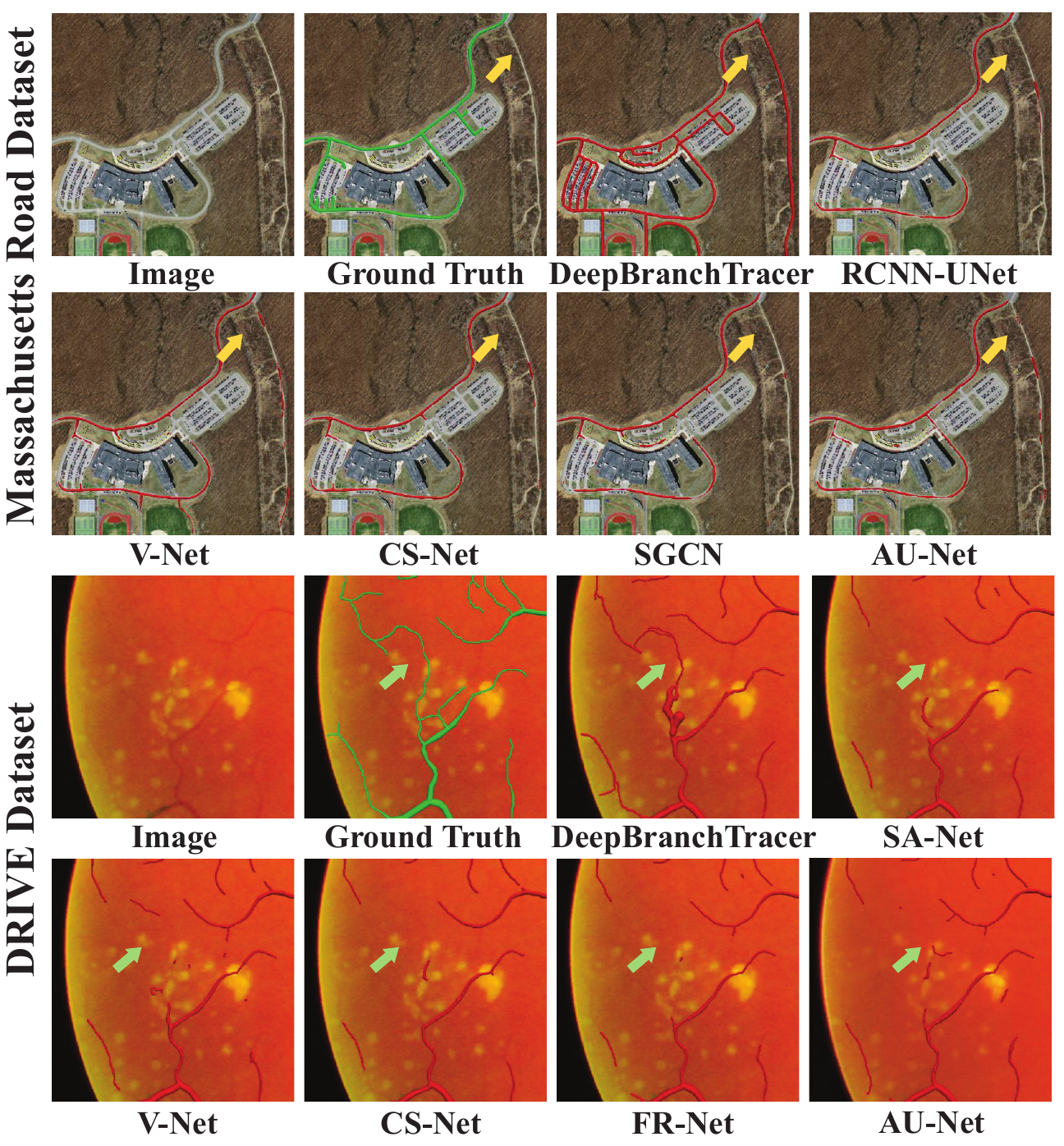}
	\caption{Reconstruction results of road and vessel datasets. The yellow arrows indicate our model detects the road which is \emph{missing} from the ground truth label. The green arrows indicate the branches are always \emph{broken} here.}
	\label{sota_figure}
\end{figure}

\begin{table}[t]
    \centering
    \small
    \setlength{\tabcolsep}{2.5pt}

    \begin{tabular}{c|c|ccccc}
    \Xhline{1.2pt}                                     
     & Model          & SSD-F1 & Precision & Recall & Length-F1 & ABL\\
    \Xhline{0.8pt}
    \multirow{8}{*}{\rotatebox{90}{Roads Dataset}} & U-Net    & 0.691  & 0.898     & 0.711  & 0.794 &  147.6  \\
     & V-Net                   & \textbf{0.703}  & 0.888     & 0.733  & 0.803 & 148.0    \\
     & AU-Net         & 0.700  & 0.899    & 0.723  & 0.801  & 150.4   \\
     & CS-Net                  & 0.680  & \textbf{0.901}   & 0.700  & 0.788 & 142.9   \\
     & RCNN-UNet               & 0.681  & 0.864     & 0.710  & 0.780  &  134.3   \\
     & SGCN                    & 0.674  & 0.872     & 0.709  & 0.782  &  140.4   \\
     \cline{2-7} 
     & RoadTracer              & 0.313   &  0.559    & 0.510   &  0.534 & 56.6 \\
     & Vec-road                & 0.635   & 0.767     & 0.805   & 0.786  & 99.8 \\
     & DBT        & 0.680  & 0.819     & \textbf{0.812}  & \textbf{0.816} & \textbf{200.2}   \\
    \Xhline{1.2pt}
    \end{tabular}
    \caption{Quantitative reconstruction results for road dataset. RoadTracer, Vec-road and DeepBranchTracer (DBT) are reconstruction-based models, while others are segmentation-based models.}
    \label{road_table}
\end{table}

\begin{figure*}[t]
	\centering
    \includegraphics[width=0.94\textwidth]{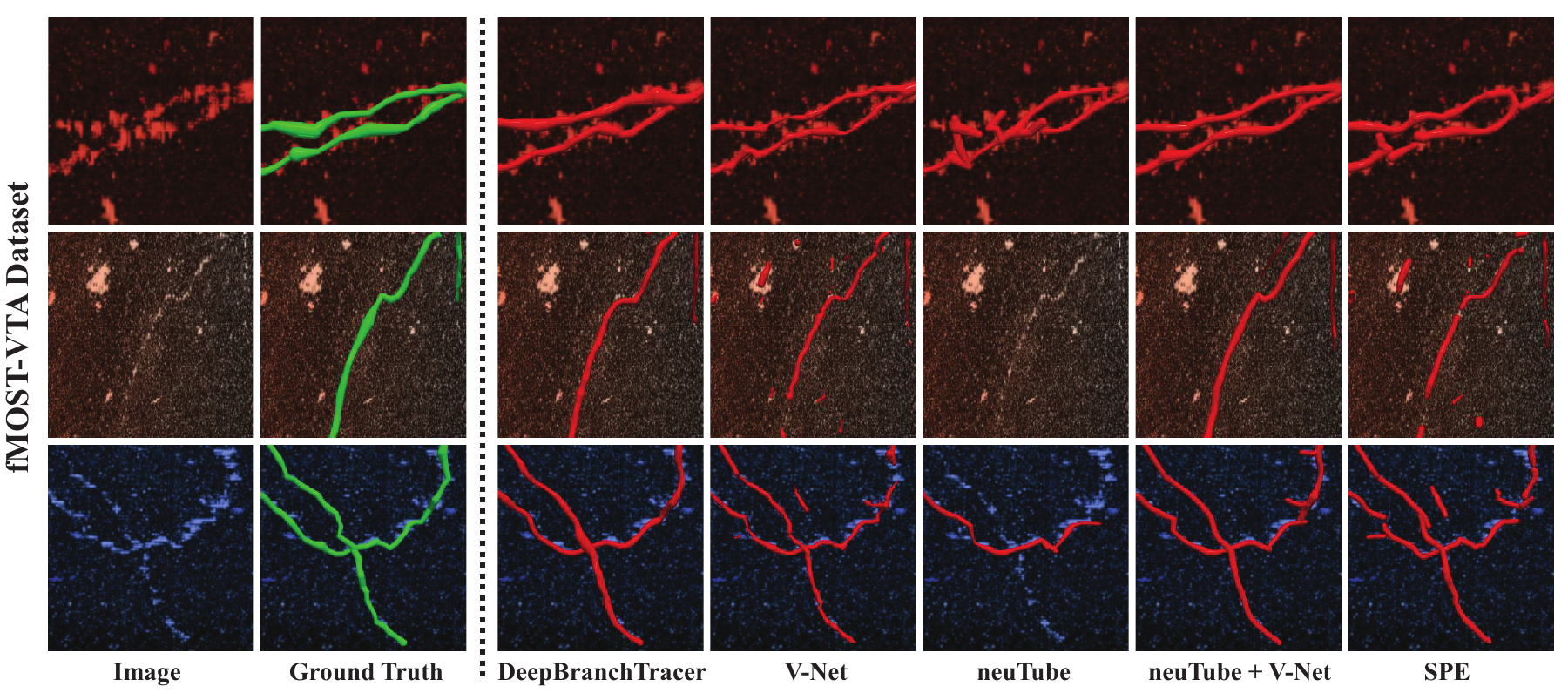}
	\caption{Reconstruction results of 3D neuron datasets. From top to bottom, we test three challenging scenarios for neuron tracing, including \emph{adjacent} branches, \emph{weak} branch, and \emph{crossover} branches.}
	\label{sota_figure_neuron}
\end{figure*}

\begin{table}[t]
    \centering
    \small
    \setlength{\tabcolsep}{4pt}

    \begin{tabular}{c|c|ccccc}
    \Xhline{1.2pt} 
      & Model &  SSD-F1 & Precision & Recall & Length-F1 & ABL \\
    \Xhline{0.8pt}
    \multirow{7}{*}{\rotatebox{90}{DRIVE}} & U-Net             & 0.841 & 0.870 & 0.768 & 0.816 & 44.5  \\
     & V-Net             & 0.847 & 0.870 & 0.773 & 0.819 & 44.4\\
     & AU-Net     & 0.847 & 0.865 & 0.772 & 0.815 & 44.6\\
     & CS-Net            & 0.847 & 0.877 & 0.774 & 0.822 & 44.0\\
     & FR-Net            & 0.857 & \textbf{0.885} & 0.784 & \textbf{0.831} & 45.4\\
     & SA-Net            & 0.856 & 0.874 & 0.784 & 0.827  & 44.3\\
     & DBT  & \textbf{0.893} & 0.827 & \textbf{0.829} & 0.828 & \textbf{67.7} \\
     \Xhline{0.8pt}
     \multirow{7}{*}{\rotatebox{90}{CHASE\_DB1}} & U-Net             & 0.774 & 0.921 & 0.757 & 0.831 & 92.6\\
     & V-Net             & 0.772 & 0.909 & 0.753 & 0.824 & 78.1\\
     & AU-Net     & 0.768 & \textbf{0.916} & 0.751 & 0.825 & 86.1\\
     & CS-Net            & 0.772 & 0.912 & 0.757 & 0.827 & 80.6  \\
     & FR-Net            & 0.776 & 0.897 & 0.770 & 0.829 & 90.7 \\
     & SA-Net            & 0.776 & 0.879 & 0.787 & 0.831 & 83.8     \\
     & DBT    & \textbf{0.836}  & 0.827 & \textbf{0.863} & \textbf{0.845} & \textbf{123.4}\\
    \Xhline{1.2pt} 
    \end{tabular}
    \caption{Quantitative reconstruction results for DIRVE and CHASE\_DB1 vessel datasets. DeepBranchTracer (DBT) is the only reconstruction-based models.}
    \label{Vessel_table}
\end{table}

\subsection{Ablation Study}

We analyzed the effects of different model settings on the fMOST-VTA dataset, as shown in Table~\ref{Ablation_table}. M1 only uses geometric features to estimate the direction and the radius of the branch. Not surprisingly, its SSD-F1 and Length-F1 was lower than the other method. M3 only uses the centerline feature to trace the branch and can be categorized as a segmentation-based method. This model yielded a lower PE score than the M1 and M2, indicating higher branch accuracy, which is also the advantage of the segmentation-based approaches. The Length-F1 score of M4 is similar to that of M3, but the ABL score is better than M3, indicating that the combination of geometric features and image features can improve the continuity of branches. It is worth noting that the ABL scores of M5 and M2 are higher than those of M4 and M1 respectively, demonstrating the effectiveness of sequence learning module. The best performance from M5 was achieved after combing all the modules and it outperforms other models in all metrics, as shown in Fig.~\ref{ablstudy_neuron}.

\subsection{Comparison with the SOTA Methods}

We tested our proposed model on the 2D road, vessel, and 3D neuron datasets respectively to show the performance of our method on both 2D and 3D fields.

\begin{table}[]
    \centering
    \small
    \setlength{\tabcolsep}{2pt}

    \begin{tabular}{c|c|ccccc}
    \Xhline{1.2pt} 
    
      & Model &  SSD-F1 & Precision & Recall & Length-F1 & ABL \\
    \Xhline{0.8pt}
    \multirow{7}{*}{\rotatebox{90}{fMOST-VTA}} & V-Net             & 0.883 & 0.868 & 0.868 & 0.868  & 71.1 \\
     & neuTube             & 0.680 & 0.833 & 0.552 & 0.664  & 103.8  \\
     & neuTube+V-Net       & 0.901 & 0.855 & \textbf{0.918} & 0.885 & \textbf{216.3} \\
     & MOST              & 0.385 & 0.429 & 0.255 & 0.320 & 12.6 \\
     & MOST+V-Net        & 0.745 & 0.684 & 0.779 & 0.728 & 23.9 \\
     & SPE               & 0.759 & 0.726 & 0.893 & 0.801 & 76.0 \\
     & DBT  & \textbf{0.910} & \textbf{0.913} & 0.909 & \textbf{0.911} & 198.3 \\
     \Xhline{0.8pt}
     \multirow{7}{*}{\rotatebox{90}{DIADEM-CCF}} & V-net     & 0.335 & 0.596 & 0.550 & 0.572 & 58.5
     \\
     & neuTube             & 0.315 & 0.519 & 0.424 & 0.467 & 63.2   \\
     & neuTube+V-Net       & 0.348 & 0.556 & 0.602 & 0.578 & 83.4   \\
     & MOST              & 0.209 & 0.411 & 0.218 & 0.285 & 25.6   \\
     & MOST+V-Net        & 0.274 & 0.509 & 0.339 & 0.407 & 16.3   \\
     & SPE               & 0.379 & 0.559 & 0.624 & 0.590 & 69.4    \\
     & DBT  & \textbf{0.421} & \textbf{0.638} & \textbf{0.627} & \textbf{0.632} & \textbf{95.7}    \\
    \Xhline{1.2pt} 
    \end{tabular}
    \caption{Quantitative reconstruction results for fMOST-VTA and DIADEM-CCF neuron datasets. neuTube+V-Net and MOST+V-Net are two reconstruction methods that rely on signal-enhanced images generated by the V-Net model.
    }
    \label{Neuron_table}
\end{table}

\subsubsection{SOTA Baselines} In this section, we set up several comparative experiments with several segmentation-based and reconstruction-based methods. We chose U-Net, V-Net \cite{milletari2016v}, and AU-Net \cite{schlemper2019attention} as classical deep segmentation-based methods, which can be conveniently applied to 2D and 3D data sets by replacing 2D convolution layer with 3D convolution layer. In particular, we also selected the SOTA methods designed specifically for various types of datasets for comparison. For the road dataset, we chose Vec-road \cite{tan2020vecroad}, RoadTracer, RCNN-UNet \cite{yang2019road} and SGCN as the comparison methods. Among them, RoadTracer is the only reconstruction-based methods. For vessel datasets, three other well-known vessel segmentation methods were adopted as the baselines, which are CS-Net \cite{mou2021cs2}, FR-Net \cite{liu2022full} and SA-Net \cite{guo2021sa}. For 3D neuron images, we also chose neuTube \cite{zhao2011automated}, MOST \cite{ming2013rapid}, and SPE as the comparison methods. Specifically, for neuTube and MOST, which are designed with hand craft features, we employ the segmentation results of 3D V-net to enhance the neuron fiber images, which is how deep learning is currently being applied to neuron reconstruction problems, namely neuTube+V-Net and MOST+V-Net.

\subsubsection{Road Datasets}

Table~\ref{road_table} summarizes the evaluation metrics for the reconstruction results of Massachusetts Roads Dataset. Among these models, our proposed model achieved the highest Length-F1 of 0.816 and Recall of 0.812. This demonstrates that our model extracts more roads compared to other methods, while still ensuring a satisfactory overall reconstruction result. The highest ABL score achieved by our model also indicates the improved continuity of the results. Most interestingly, our model also detects the road which is \textbf{missing} from the ground truth label, suggesting the high sensitivity of our method to curvilinear structures, as shown in the second rows of Fig.~\ref{sota_figure}. Compared with the ground truth, these newly detected roads are mistaken for noise, resulting our method has low Precision scores.

\subsubsection{Vessel Datasets}
The DRIVE and CHASE\_DB1 datasets are typically considered as segmentation tasks. In this study, we treat vessel segmentation as a reconstruction task, aiming to explore a novel perspective for addressing the challenge of extracting weak vessels. The results of reconstruction on the vessel datasets are shown in Table~\ref{Vessel_table}. Our proposed model obtained the highest SSD-F1, Recall, and ABL scores on DRIVE, and the highest SSD-F1, Recall, Length-F1 and ABL on CHASE\_DB1. It is worth noting that the Precision score of our model is low and the Recall score is the highest. We consider that the model further improves the extraction ability of weak vessels while inevitably increasing the tolerance of false positives. As shown in Fig.~\ref{sota_figure}, our model extracted more weak vessels than the other methods, and also achieved the best reconstruction effects. What's more, with similar coverage rate of branches, our model can trace longer branches, reflected in the higher ABL scores.

\subsubsection{Neuron Datasets}
The 3D neuron reconstruction results of the compared method on the fMOST-VTA and DIADEM-CCF datasets are shown in Table~\ref{Neuron_table}. It can be observed that in the fMOST-VTA image, the branch signal is always discontinuous in the microscope image due to uneven staining and optical system fluctuation, and can pose a challenge to automatic reconstruction, as shown in Fig.~\ref{sota_figure_neuron}. Among all the methods, our DeepBranchTracer achieved the superior performances, including the highest SSD-F1, Precision, Length-F1 in fMOST-VTA dataset, and the highest all metrics in DIADEM-CCF dataset. The neuTube+V-Net also achieved satisfactory ABL score in fMOST-VTA dataset. However, neuTube was designed by hand-craft features and only suitable for dark field images, limiting its applicability within certain applications. The visualization results show that in most challenging scenarios, DeepBranchTracer consistently achieved better performance, demonstrating its robustness and generalization.

\section{Conclusion}
Curvilinear structure extraction is an essential research area, but there are still challenges due to the complexity of its geometry. In this paper, we propose a novel curvilinear structure reconstruction method, namely DeepBranchTracer, which can be conveniently applied to various 2D and 3D image datasets. Specifically, our DeepBranchTracer utilizes both the external image features and internal geometric features of the curvilinear structure to finish the tracing process iteratively. Compared with the existing tracing method, our DeepBranchTracer does not require any parametric geometry model or complex training strategy, demonstrating its strong effectiveness and applicability. We evaluate our method on five road, vessel and neuron datasets, and the results show our model has a great potential in complex-topology curvilinear structure extraction.

\section{Acknowledgments}
This work is supported by the National Key R\&D Program of China (2020YFB1313500), National Natural Science Foundation of China (T2293723, 61972347), Zhejiang Provincial Natural Science Foundation (Z24F020008), the Key R\&D Program of Zhejiang Province (2022C01022, 2022C01119, 2021C03003), and the Fundamental Research Funds for the Central Universities (No. 226-2022-00051). 


\bibliography{camera-ready-AAAI2024-2240}

\clearpage

\end{document}